\documentclass[10pt,twocolumn,letterpaper]{article}

\usepackage[pagenumbers]{cvpr} 

\usepackage[ruled,noend]{algorithm2e} 

\SetAlFnt{\small}
\SetAlCapFnt{\small}
\SetAlCapNameFnt{\small}
\SetAlCapHSkip{0pt}
\IncMargin{-\parindent}
\usepackage{ifpdf}
\usepackage{graphicx}

\usepackage{color}
\definecolor{cyan}{cmyk}{1,0,0,0}
\definecolor{darkgreen}{rgb}{0,0.5,0}
\definecolor{orange}{rgb}{1,0.5,0}
\definecolor{magenta}{cmyk}{0,1,0,0}
\definecolor{darkyellow}{cmyk}{0,0,0.75,0}
\definecolor{gray}{rgb}{0.8,0.8,0.8}

\usepackage{amssymb} 
\usepackage{amsmath} 

\usepackage[toc,page]{appendix} 
\usepackage{wrapfig} 
\usepackage{comment}
\usepackage{bm}
\usepackage{cuted} 
\usepackage{lipsum} 
\usepackage{mathtools} 
\usepackage{algorithmicx}
\usepackage{algpseudocode}
\usepackage{nicefrac}

\usepackage{gensymb}
\usepackage{float}
\usepackage{soul}
\usepackage{array}
\usepackage{multirow}
\usepackage{hhline}
\usepackage{setspace}
\usepackage{nicefrac}

\algdef{SE}[DOWHILE]{Do}{doWhile}{\algorithmicdo}[1]{\algorithmicwhile\ #1}%

\makeatletter
\renewcommand{\ALG@beginalgorithmic}{\small}
\makeatother

\newcommand{\DELETE}[1]{} 
\newcommand{\IGNORE}[1]{}

\usepackage{graphicx}
\usepackage{amsmath}
\usepackage{amssymb}
\usepackage{booktabs}
\usepackage[table]{xcolor}
\usepackage[symbol]{footmisc}

\usepackage[pagebackref,breaklinks,colorlinks]{hyperref}

\def\cvprPaperID{6200} 
\def\confName{CVPR}
\def\confYear{2022}
\definecolor{red}{rgb}{1,0,0}

\newif\ifthesis
\thesisfalse

\begin{document}

\title{\vspace{-2mm}Neural Ray-Tracing: \vspace{5pt}\\Learning Surfaces and Reflectance for Relighting and View Synthesis\vspace{-5mm}}
\author{
  Julian Knodt \qquad
  Joe Bartusek\footnotemark[2] \qquad
  Seung-Hwan Baek \qquad
  Felix Heide
  \vspace{1mm}\\
  Princeton University
	\vspace{2mm}
}

\twocolumn[{%
\vspace{-7mm}
\renewcommand\twocolumn[1][]{#1}%
\maketitle
\vspace{-3mm}
\thispagestyle{empty}
}]

\begin{abstract}
Recent neural rendering methods have demonstrated photo-realistic view interpolation by predicting
volumetric density and color with a neural network. Although such volumetric representations can
be supervised on static and dynamic scenes, existing methods implicitly bake the scene
light transport into a single neural network for a given scene, entangling surface modeling,
bidirectional scattering distribution functions, and indirect lighting effects. In contrast to
traditional rendering pipelines, this prohibit decomposing scene transport in reflectance and geometry and changing illumination conditions later.

In this work, we explicitly model the light transport between scene surfaces by proposing disentangled neural representations of geometry and reflectance, allowing for efficient inverse rendering.
Furthermore, we present a disentangled rendering method inspired by a standard ray-tracing pipeline, inheriting the rendering interpretability for multi-bounce reflection.
Combining the neural scene representations and rendering method enables efficient and effective inversion of the rendering equation for inverse rendering.
We quantitatively and qualitatively evaluate our method and confirm our approach outperforms existing methods for relighting and view synthesis. Find code under~\url{https://github.com/princeton-computational-imaging/neural_raytracing}.
\end{abstract}

\footnotetext[2]{now at Columbia University.}

\begin{figure}
\centering
\vspace{2mm}
  \includegraphics[width=\linewidth]{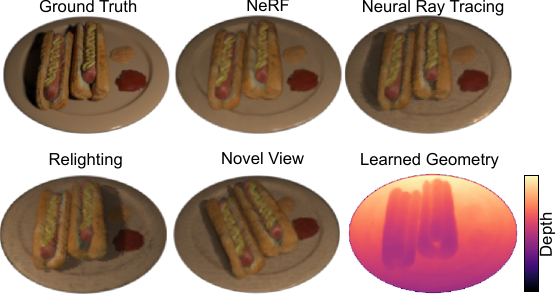}
  \caption{
    The proposed neural ray-tracing method relies on a novel learned diffuse and non-diffuse
    BRDF model in conjunction with a SDF-based implicit surface model~\cite{yariv2021volume}.
    This allows us to formulate a path-tracing differentiable rendering pipeline, including
    raycasted occlusion. Allowing for decomposable  neural ray tracing method is able to recover
    accurate surface normals as compared to NeRF~\cite{mildenhall2020nerf}. In addition, NeRF
    cannot capture any sort of occlusion, whereas our model is able to accurately predict
    shadows.
    \vspace{-2mm}
  }
\end{figure}

\section{Introduction}

View synthesis and scene reconstruction from a series of images is a fundamental problem in
computer vision and graphics.
Conventional ray-tracing approaches model physical-light transport, decomposing it into illumination, reflectance, and multi-path scene inter-reflections.
These representations are interpretable and offer great editability required for many visual-computing applications such as view synthesis and relighting.
However, directly inverting the rendering equation is not trivial in ray-tracing.
While remarkable progress has been recently made~\cite{nimier2020radiative,Zhang:2020:PSDR,nimier2020radiative}, bypassing the non-differentiable representations and operations in ray tracing is still challenging, limiting its applicability to complex scenes.


Recent advances in neural scene rendering have achieved photo-realistic view synthesis results by encoding the entire scene light transport into a learned volumetric representation~\cite{flynn2019deepview,mildenhall2020nerf,yariv2020multiview,mildenhall2019llff}.
Here, the representations and the rendering procedure are trivially differentiable allowing for modeling complex real-world scenes.
However, all light transport components are baked into a higher-dimensional radiance field: because of the implicit radiance representations that these works rely on, light transport, surface representations, and scattering interactions are entangled.
This prevents any single component from being recovered or modified without fundamentally altering the others, prohibiting relighting.
Recently, attempts to incorporate relighting into existing neural pipelines have been
proposed~\cite{bi2020neural, bi2020deep, srinivasan2020nerv}. While these approaches allow for
relighting, they rely on dense volumetric sampling for lighting and analysis which limits its variability and rendering efficiency~\cite{zhang2021nerfactor}.

In this work, we combine the benefits of physically-based ray-tracing and neural scene representations by using neural representations in the middle of a ray-tracing rendering pipeline to disentangle shape, reflectance, and lighting. This enables decomposing the baked scene information into each component and hence editing them to accurately relight and simulate a novel view. This provides a bridge between the recent physically-based differentiable rendering~\cite{nimier2020radiative,Zhang:2020:PSDR,nimier2020radiative} and volumetric neural rendering methods~\cite{flynn2019deepview,mildenhall2020nerf,yariv2020multiview,mildenhall2019llff}

The proposed method exploits neural volumetric geometry model as a signed-distance
fields~\cite{yariv2021volume} and a novel disentangled reflectance representation for describing a spatially-varying bidirectional scattering distribution function (BSDF).
We then present a disentangled rendering method which utilizes the neural representations.
Specifically, we partition the optimization space of into the discrete problems of
surface estimation, and reflectance estimation by performing physically-based integration and ray-surface intersection during evaluation.
This naturally inherits the benefits of physically-based ray-tracing methods for multi-bounce rendering, allowing for efficient simulation of multi-bounce reflection to $\mathcal{O}(n\log{n})$ from $\mathcal{O}(n^2)$ complexity.
We can also interpret our method as incorporating emerging neural scene representations using multi-layer perceptrons (MLP) in a physically-based rendering framework in order to achieve the universality of neural
representations~\cite{mildenhall2020nerf,bi2020neural,yariv2020multiview} and the interpretability of physically-based ray tracing methods. 
We extensively evaluate the proposed method and demonstrate that we outperform existing methods.
In this paper, we make the following contributions:
\begin{itemize}
\itemsep0em
\item We present an end-to-end forward and inverse rendering approach with neural scene
reconstructions and ray tracing, allowing for decomposing shape and reflectance in consideration of multi-bounce transport.
\item We propose a novel decomposed BSDF model along with efficient path-tracing
intersection computation that allows us to recover higher-order bounces quickly.
\item The proposed method outperforms existing neural rendering methods qualitatively and quantitatively.
\end{itemize}

Although the proposed method could be extended to unknown lighting conditions, the scope of this work is
known co-located lighting conditions, see also~\cite{srinivasan2020nerv}, to allow for minimal
ambiguity between reflectance and lighting conditions.

\section{Related Work}
\subsection{Monolithic Scene Representation}
With the advent of deep learning and neural networks, a growing body of work has explored
methods for scene representations with convolutional neural networks~\cite{lombardi2019neural,
park2019deepsdf,sitzmann2019deepvoxels} and implicit neural
networks~\cite{mescheder2019occupancy,peng2020convolutional,mildenhall2020nerf}. Notably,
modeling the scene radiance with a neural implicit function, commonly in the form of a MLP, has
proven to be effective, as demonstrated in NeRF~\cite{mildenhall2020nerf}. MLP-based scene
modeling integrates light radiance in a volumetric structure and optimizes the volumetric
radiance function as a MLP that accurately reproduces a set of reference images, demonstrating
learned view-dependent effects. While previous works are effective at novel view synthesis,
most of them entangle reflectance, shape, and lighting in a single radiance volume, limiting its
usage for physically-accurate scene decomposition~\cite{mildenhall2020nerf,yariv2020multiview}.
In contrast, we learn physically-based scene representations of shape and reflectance. This enables high-quality, efficient rendering with interpretability, removing black boxes in NeRF.

\subsection{Geometry-aware Scene Reconstruction}
Structure-from-Motion (SfM) has been a golden-standard of reconstructing scene geometry, given a series of images captured at different camera positions~\cite{snavely2006photo,schoenberger2016sfm}, however directly using it for synthesizing novel views and relighting is challenging given the sparsity of reconstruction from SfM methods.
Recently, implicit differentiable rendering demonstrates effective shape reconstruction based on volumetric signed distance fields~\cite{yariv2020multiview}.
Zhang et al.~\cite{zhang2021physg} builds on this geometric representation and employ a spherical Gaussian reflectance model. However, it does not support multi-bounce rendering, which is the core of many physically-based rendering methods.
Our work is based on a neural volumetric model, however while previous methods model the entire
rendering equation as a black-box neural network, we replace this with explicit integration, inspired by ray tracing to better understand and structure each scene component of geometry, reflectance, and lighting.

\subsection{Reflectance-aware Scene Representation}
With the goal of extending radiance to reflectance, recent works have also explored decomposed
modelling of reflectance and volumetric density to enable
editable materials and lighting~\cite{bi2020neural,srinivasan2020nerv}. These works often rely on
approximations to volumetric rendering, such as depth-termination prediction. It additionally
makes multi-bounce rendering computationally expensive (e.g., 128 TPUs for training~\cite{srinivasan2020nerv}).
Zhang et al.~\cite{zhang2021nerfactor} proposed a neural rendering method to decompose the baked NeRF representation into normals, albedo, and illumination given an unknown illumination condition.
All previous work relies on volumetric rendering, demonstrating the lack of representability of
discrete structures used in forward rendering pipelines in traditional graphics.
Our work attempts to bridge the gap between the successes of neural scene representations for volumetric inverse rendering and physically-based ray-tracing methods for interpretable high-fidelity rendering capability.


\subsection{Physically-based Differentiable Ray-Tracing}
Ray-tracing has been extensively studied with the goal of generating photorealistic
imagery using the rendering equation~\cite{PBRT3e}. Inverting the rendering equation is used to
estimate scene compositions and parameters from a given set of input images.
Recently, differentiable path-tracing has gained significant interest through frameworks implementing the forward rendering operations in a differentiable manner, so that backpropagation can be applied to optimize over some loss over
the rendering pipeline~\cite{NimierDavidVicini2019Mitsuba2,nimier2020radiative,zhang2020path}. While these
efforts demonstrate promising results, this inversion still faces great challenges. One of the central
problems is the conventional representation of shape and reflectance. Shapes are commonly
represented as a mesh in existing rendering approaches, imposing non-differentiable geometric
discontinuity. While different approaches have been proposed to tackle this
problem~\cite{henderson2020leveraging}, it is still an open challenge to effectively and
efficiently invert transport using the previous scene representations often used for forward rendering. Similarly, analytic
reflectance models do not facilitate effective learning via gradient-based optimization, often due to
discontinuities over lighting or low-likelihood of sampling specular highlights.
In this work, we aim to inherit the benefits of physically-based ray tracing by using its integration and sampling schemes and the differentiable neural scene representations of shape and reflectance.

\begin{figure}[t]
  \centering
  \includegraphics[width=\linewidth]{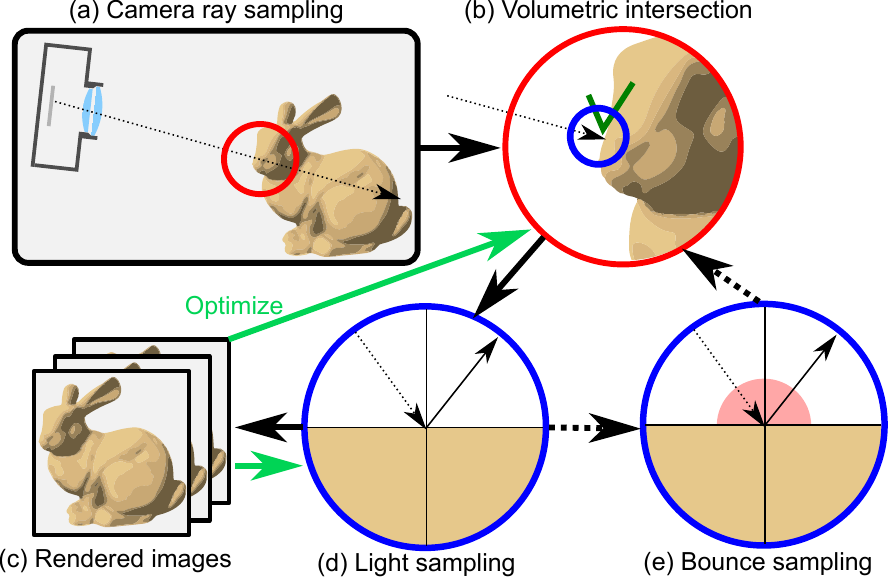}
  \caption{
    Overview of Neural Ray-Tracing's pipeline. We ray-march for each pixel in an image,
    computing the normals of the intersection point with the predicted SDF. Then, we sample the
    direction of the incident light and obtain the reflected energy using its BRDF value. This
    process is repeated for path tracing, enabling rendering with multiple bounces. Finally, the
    pixel value is computed using volumetric integration over the RGB at each ray-marched point.
  }
  \vspace{-3mm}
\end{figure}

\section{Neural Ray-Tracing}
The rendering equation proposed by Kajiya~\cite{Kajiya86therendering} has served as the
foundation for photo-realistic rendering, modeling both direct and indirect light transport.
Specifically, the rendering equation describes the outgoing radiance $L_o$ coming from a scene point
$\mathbf{x}$ toward a direction $\omega_o$ as
\begin{align}\label{eqn:rendering}
L_o(\mathbf{x},\omega_o) = &L_e(\mathbf{x},\omega_o) + \nonumber \\ &\int_{\Omega}
f(\mathbf{x},\omega_i,\omega_o) L_i(\mathbf{x},\omega_i) (\omega_i\cdot \mathbf{n}) d\omega_i,
\end{align}
where $L_e$ and $L_i$ are the emitted and incident light radiance, $\omega_i$ is an incident
direction, $f$ is the BRDF, and $\mathbf{n}$ is the surface normals.
All the indirect contribution from other scene points are integrated over the hemisphere
$\Omega$.
Solving this rendering equation is commonly referred as forward rendering and it has been
extensively studied in computer graphics. Existing forward rendering methods compute the integral from Eq.~\eqref{eqn:rendering} by pathtracing using meshes, analytic reflectance and lighting models as scene representation~\cite{PBRT3e}.
While being effective for forward rendering, this approach does not lend itself towards solving the inverse problem: estimating scene parameters from observed measurements~\cite{NimierDavidVicini2019Mitsuba2}.
Instead, recent neural methods instead take a volumetric rendering approach that bakes all scene light transport into a 5D radiance field~\cite{mildenhall2020nerf}.
A priori, This prohibits modeling indirect reflection, and complicated interplay between geometry, reflectance, and lighting.

Instead of relying on one of the extremes, we propose neural ray-tracing method which attains the rich light simulation properties of path tracing while resolving the non-differentiable nature of conventional
representations by using fully-differentiable neural representations of surfaces and
reflectance.
Our method defines a differentiable forward rendering function $\mathrm{Render}()$ which takes
the neural representation of shape and reflectance, allowing for iteratively optimizing them
with a first-order optimization framework.
We start by describing the neural representation of shape and reflectance.

\paragraph{Disentangled Reflectance}
Modeling material reflectance by BSDF, $f$, as in Eq.~\eqref{eqn:rendering} has been extensively
studied in computer graphics and vision. Analytical models such as the Phong
BSDF~\cite{Phong1975Shading} and microfacet BSDFs~\cite{walter2007microfacet} have provided an
efficient way of representing real-world reflectance with few parameters.  Recently, researchers
have proposed neural representations of reflectance that provide differentiability and high
representation power with more parameters compared to their analytical
counterparts~\cite{hu2020deepbrdf,chen2020invertible}. However, learned models do not provide
strong priors on reflectance, therefore directly using it for inverse rendering and accurate
recovery is challenging and inefficient. We combine the benefits of both analytical and learned
BSDF models by representing reflectance with an analytic Lambertian component and a neural
components for the residual learning including specular highlights.  This not only maintains the
strong representation power of neural representations, but also allows for efficient
reconstruction thanks to the parameter efficient Lambertian model. While our model is able to
predict albedo in the range $[-1,2]$, a range that includes non-energy preserving values outside 
of $[0,1]$, it is trained on images in the range $[0,1]$, and the diffuse model provides sufficient initialization to prevent converge to outside that range. In practice, we do not find the energy preservation to affect the reconstruction quality. 

\begin{figure}
  \centering
  \includegraphics[width=\linewidth]{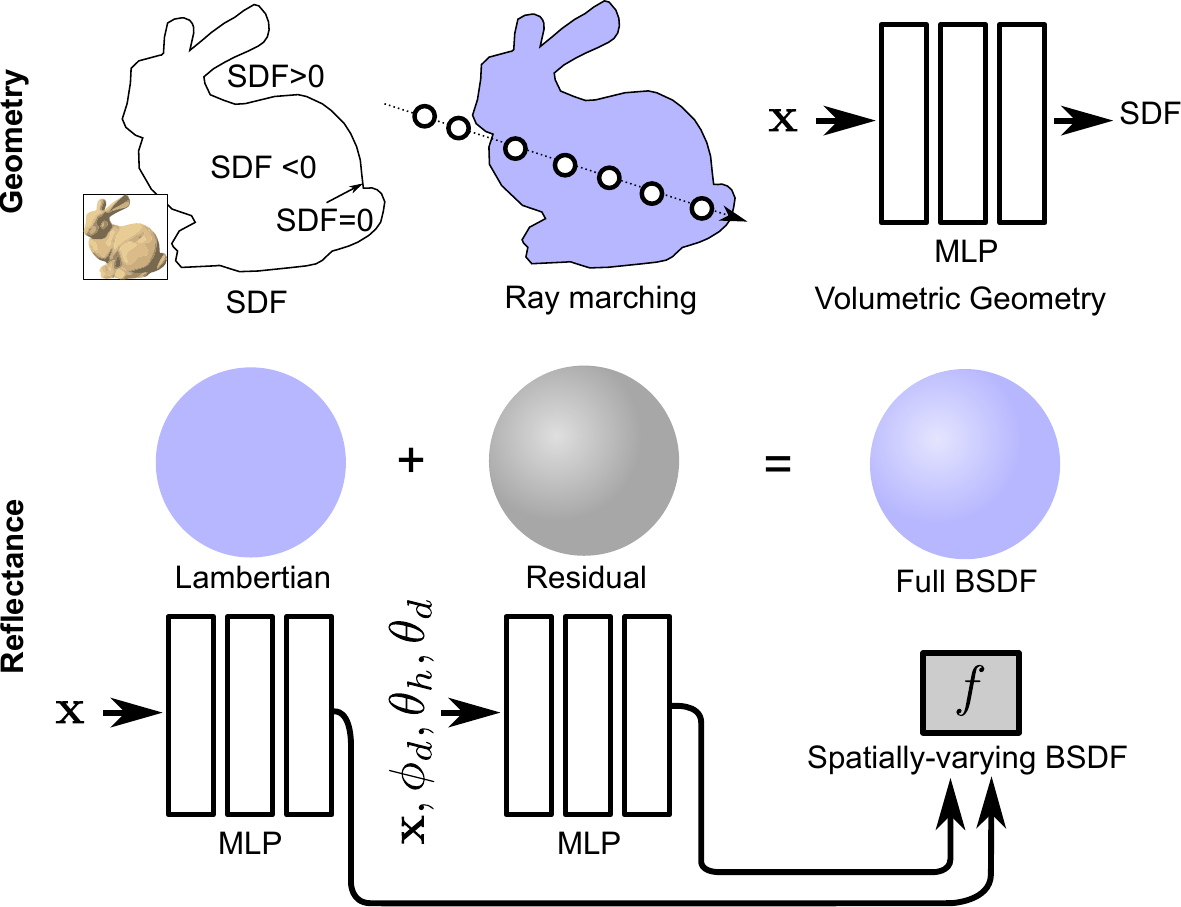}
  \caption{
    \label{fig:representation}
    Proposed Neural Scene Representation: Using an SDF modeled as an MLP, we can render
    high-quality geometry using volumetric integration for primary rays and ray-marching
    with detached gradients for finding intersections for memory-efficient secondary bounces.
    For the reflectance, we compose a diffuse Lambertian BSDF with a learned BSDF in order to
    represent a high-quality spatially-varying BSDF. We combine a learned BSDF with a fully
    diffuse model in order to enforce correct normals during reconstruction.
  }
  \vspace{-3mm}
\end{figure}
Specifically, the Lambertian model component is described as a single diffuse albedo
$k_d(\mathbf{x})$, modeling foreshortened diffuse reflection $k_d(\mathbf{x})(\omega_i \cdot
\mathbf{n})$.  We note that this model component also indirectly places a \emph{strong prior on
the normals}, which is necessary for accurate reconstruction. We model the diffuse color
$k_d(x)\in[0,1]$ as a Fourier MLP~\cite{tancik2020fourfeat} with a sigmoid activation to
constrain the value to be positive and energy-preserving.
For the learned BSDF component, we introduce an MLP that takes incident and exitant angles with
respect to the surface normals and outputs trichromatic reflectance as
$\tanh\left(\mathrm{MLP}(\mathbf{x}, \phi_d,\cos{\theta_h},\cos{\theta_d})\right)$, where
$\phi_d$, $\theta_h$, $\theta_d$ are the Rusinkiewicz angles~\cite{rusinkiewicz1998new}.
The Rusinkiewicz angles provide effective angular information and has been proven to be effective in microfacet BSDF models~\cite{walter2007microfacet}.
Note that we use the hyperbolic tangent function for activation to allow for both negative and
positive values.
This is crucial because residual error of the Lambertian model could be found in any direction.
In summary, the proposed reflectance model, illustrated in Figure~\ref{fig:representation}, is defined as follows
\begin{align}
\label{eq:bsdf}
  f(\mathbf{x}, \omega_o, \omega_i) = f_\mathrm{Lambert}(\mathbf{x}) + f_\mathrm{Learned}(\mathbf{x}, \phi_d,\theta_h,\theta_d),
\end{align} %
where $f_\mathrm{Lambert}(\mathbf{x})$ is the Lambertian BSDF defined as
$\sigma\left(\mathrm{MLP}(\mathbf{x})\right)$. $\sigma$ is the sigmoid function.
The function $f_\mathrm{Learned}(\mathbf{x}, \phi_d,\theta_h,\theta_d)$ is the learned residual BSDF defined as $\tanh\left(\mathrm{MLP}(\mathbf{x}, \phi_d,\cos{\theta_h},\cos{\theta_d})\right)/(\omega_i \cdot \mathbf{n})$.

\subsection{Disentangled Neural Rendering}
With our neural representations of geometry and reflectance, we now define the forward rendering
function $\mathrm{Render}()$ and describe how to optimize the scene parameters through the
rendering function. Ultimately, the goal is to combine pathtracing and
neural representations, which imposes a challenge in computational efficiency. Pathtracing
requires multiple samples of indirect light path resulting in an exponentially number of MLP
evaluations with more bounces.  In order to resolve this problem, we use a disentangled
alternating rendering approach which allows for effective optimization.

\paragraph{Direct Reflection and Occlusion Learning}
To start the training procedure, we start by focusing on direct reflection and learn the geometric SDF and the Lambertian BSDF: $\mathrm{SDF}(\mathbf{x}) = \mathrm{MLP}(\mathbf{x})$ and $f=f_\mathrm{Lambert}(\mathbf{x})$.  Specifically, we define the forward
rendering function by casting rays toward a camera pixel $x$ and integrate along the ray using
the geometric SDF. For each point $\mathbf{x}$ along the ray, we sample the direct paths
$\omega'$ to the light sources with known positions.  The visibility of the point $\mathbf{x}$ from
the light position is modeled as an MLP with sigmoid activation as
$O(\mathbf{x},\omega')=\sigma(\mathrm{MLP}(\mathbf{x},\omega'))$.
We then compute the radiance $p$ coming from the point $\mathbf{x}$ toward the pixel $x$ as
\begin{equation}\label{eq:radiance1}
p(\mathbf{x}) = \sum\limits_{\mathbf{l} \in {L}} (\mathbf{n} \cdot \dot{\overrightarrow{\mathbf{lx}}}) f({\bf{x}},{\omega _o},\dot{\overrightarrow{\mathbf{lx}}})O({\bf{x}},\dot{\overrightarrow{\mathbf{xl}}})E_\mathbf{l},
\end{equation}
where $E_{\mathbf{l}}$ is the radiance for the illumination $\mathbf{l}$. Here, $L$ is the set of light sources, and $\dot{\overrightarrow{\mathbf{lx}}}$ is the unit vector from the light source to the point.
Given the radiance $p$, we can render the pixel value $I(x)$ using volumetric integration
\begin{equation}\label{eq:forward1}
{I(x)} = \int T (t){\Psi }({\mathrm{SDF}}(\mathbf{x}(t)))p(\mathbf{x}(t))dt,
\end{equation}
where $\mathbf{x}(t)$ is the sampled point along the emitted ray from the camera pixel $x$ as $\mathbf{x}(t) = \mathbf{o} + t \dot{\overrightarrow{\mathbf{o}x}}$. The function $T$ is the transparency function defined as $T(t) = \exp ( - \int_{{0}}^t {{\Psi }} ({\rm{SDF}}(\mathbf{x}(t))){\rm{dt}})$. Here, $\Psi$ is the volumetric equivalence of an SDF, that is
${\Psi}(s) = \frac{1}{2}\exp (\frac{s}{\beta})$ if $s \le 0$, otherwise ${\Psi}(s)=1 - \frac{1}{2}\exp ( - \frac{s}{\beta })$.
We assume here that the camera and the illumination are known. The proposed method relies on this forward rendering function to learn
the neural scene representations of the geometric SDF, the Lambertian BSDF, and the occlusion
function by solving the following optimization problem
  \begin{align}\label{eq:inverse1}
    \mathop {\mathrm{min}}\limits_{\mathrm{SDF}, f, O} &\|
    I(x;\mathrm{SDF}, f, O)- \mathrm{I}_\mathrm{ref}(x) \|_2 \nonumber + R(\mathrm{SDF}),
  \end{align}
where $R$ is the eikonal regularization~\cite{gropp2020implicit} on the surface, that is $R(\mathrm{SDF}) = \mathbb{E}_{\forall \mathbf{x}}[||\nabla\mathrm{SDF}(\mathbf{x})||-1]^2$.
Note that the optimization above is performed jointly over over all camera views and light sources.

\paragraph{Residual Reflectance and Geometry-aware Occlusion}
In the next optimization step, we increase the model capacity by introducing learned reflectance and geometry-aware occlusion terms.
Specifically, we expand the Lambertian-only reflectance term to the full representation in Equation~\eqref{eq:bsdf} while maintaining the previously-learned Lambertian MLP by only adding the residual MLP with the tangent activation. Next, we increase the capacity of the occlusion model. In the previous stage, our occlusion function $O$ outputs the value from zero and one with the sigmoid activation, however, there is no guarantee that the learned occlusion $O$ provides occlusion effect corresponding to the learned geometric SDF.
To solve this problem, we first efficiently compute the visibility of the point $\mathbf{x}$ with respect to the light direction $\omega'$ by ray casting on the fixed geometry, resulting in a boolean indicator function $\mathbb{I}_\text{visible}(\mathbf{x},\omega')$.
We then expand the occlusion term as
\begin{small}
\begin{align}
  O(\mathbf{x},\omega')=\sigma(\mathrm{MLP}(\mathbf{x},\omega'))
  (\mathbb{I}_\text{v}(\mathbf{x},\omega')+\sigma(k)(1-\mathbb{I}_\text{v}(\mathbf{x},\omega'))),
\end{align}
\end{small}
where $k$ is a scalar that determines the softness of the raycast shadow.
The first term $\sigma(\mathrm{MLP}(\mathbf{x},\omega'))$ is here the pretrained model in the first stage.
The second term utilizes the visibility function obtained from the fixed geometry and allows for soft shadows.
This enables geometric-dependent occlusion modeling.
With the updated reflectance and occlusion terms, the inverse procedure then optimizes the learned reflectance and the scalar $k$ as follows
\begin{align}\label{eq:inverse2}
  \mathop {\mathrm{minimize}}\limits_{f, O} &\| I(x;\mathrm{SDF}, f, O)- I_\mathrm{ref}(x) \|_2.
\end{align}

\begin{figure}[t]
  \centering
  \includegraphics[width=\linewidth]{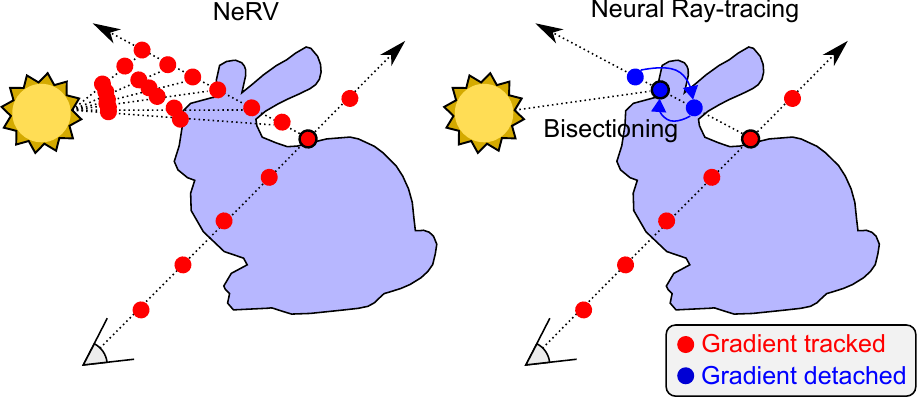}
  \caption{
  \label{fig:nerv_comparison_pathtracing}
    During training, NeRV~\cite{srinivasan2020nerv} performs volumetric sampling for both primary and secondary rays to account for multi-bounce reflection.
    In contrast, our method learns an occlusion MLP at every point given a direction, allowing
    for efficient training. We fine-tune direct integration with raycasting with detached
    gradients, and for pathtracing we perform additional bisectioning to find precise
    intersection points.
  }
  \vspace{-2mm}
\end{figure}
\paragraph{Multi-bounce Path Tracing}
As the next update step, we incorporate multi-bounce reflections via path tracing by exploiting the previously optimized scene representations.
Specifically, for every point $\mathbf{x}$ along the initial ray, we sample directions on the hemisphere.
For each sampled direction $\omega'$, we find the sign-change region of the optimized SDF which indicates the existence of surface. We then compute the accurate surface intersection $\mathbf{x}'$ by binary searching the zero set of the SDF within that range.
Computational complexity reduces to $O(N \log N)$ compared to $O(N^2)$ of NeRV~\cite{srinivasan2020nerv}.
Furthermore, our approach is gradient-computation free, resulting in computational efficiency and less memory usage than volumetric integration.
See Figure~\ref{fig:nerv_comparison_pathtracing}.
We can then represent the second-bounce contribution to the pixel by accumulating the BSDF evaluations on the original point $\mathbf{x}$ and the intersection point $\mathbf{x}'$ as $f(\mathbf{x},\dot{\overrightarrow{\mathbf{x}x}},\dot{\overrightarrow{\mathbf{x}'\mathbf{x}}}) \times f(\mathbf{x}', \dot{\overrightarrow{\mathbf{x}'\mathbf{x}}}, \dot{\overrightarrow{\mathbf{l}\mathbf{x}'}})$.
In this transport estimate, we also include an additional transfer function $G$ which approximates the joint distribution of the points $\mathbf{x}$ and $\mathbf{x}'$ as a MLP with sigmoid activation.
That is, the radiance from the point $\mathbf{x}$ is defined as
\begin{small}
\begin{align}\label{eq:forward3}
  &{p}(\mathbf{x}) = \sum_{\mathbf{l}\in L} \{\underbrace{(\mathbf{n} \cdot
  \dot{\overrightarrow{\mathbf{lx}}}) f({\bf{x}},{\omega
  _o},\dot{\overrightarrow{\mathbf{lx}}})O({\bf{x}},\dot{\overrightarrow{\mathbf{xl}}})E_\mathbf{l}}_{\mathrm{Direct}}
  \nonumber \\ &+ \underbrace{\sum_{\mathbf{x}'} (\mathbf{n} \cdot
  \dot{\overrightarrow{\mathbf{x}'\mathbf{x}}})(\mathbf{n}' \cdot
  \dot{\overrightarrow{\mathbf{lx}'}})f(\mathbf{x},\dot{\overrightarrow{\mathbf{x}x}},\dot{\overrightarrow{\mathbf{x}'\mathbf{x}}})
  f(\mathbf{x}', \dot{\overrightarrow{\mathbf{x}'\mathbf{x}}},
  \dot{\overrightarrow{\mathbf{l}\mathbf{x}'}})G(\mathbf{x},\mathbf{x}')E_{\mathbf{l}}}_{\mathrm{Indirect}}\},
\end{align}
\end{small}
where $\mathbf{n}'$ is the surface normals at the point $\mathbf{x}'$.
Obtaining the pixel intensity is performed by volumetric integration of Equation~\eqref{eq:forward1}.
We solve the corresponding inverse problem by optimizing the reflectance, occlusion parameters, and the transfer function as
\begin{align}\label{eq:inverse3}
  \mathop {\mathrm{minimize}}\limits_{f, O, G} &\| I(x;\mathrm{SDF}, f, O)- I_\mathrm{ref}(x) \|_2.
\end{align}

\paragraph{Training}
We alternately solve the individual inverse problems in Eqs (\ref{eq:inverse1},\ref{eq:inverse2},\ref{eq:inverse3})
using the Adam optimizer for all inverse problems.
We set the learning rate to $3\times10^{-4}$, with cosine annealing to
a minimum of $5\times10^{-5}$. All hyperparameter settings are specified in the Supplemental
Document, and a precise specification can be found in the implementation.

We train on the two public datasets from NeRV~\cite{srinivasan2020nerv} with single point lights
which contain non-collocated point light sources, and also compile our own dataset using a
subset of the original NeRF scenes, but with collocated point light sources at every view.
Collocated point light sources have the benefit that they allow for efficient reconstruction
since there are minimal shadows so there is minimal ambiguity between BSDF and lighting
conditions. Our method can also reconstruct non-collocated conditions, by introducing shadows
using a combination of a learned approximation and raycasting.
Our dataset re-renders new configurations of assets used in NeRF~\cite{mildenhall2020nerf},
consisting of 7 scenes with collocated point lights which will be made available upon acceptance.

\section{Assessment}
We assess the proposed model by analyzing the light transport decompositions that our approach
learns, and how plausible it is for relighting. We train and evaluate on datasets with known illumination, that is the dataset proposed in~\cite{srinivasan2020nerv} and a collocated synthetic dataset rendered with collocated illumination on the same synthetic scenes as in~\cite{mildenhall2020nerf}.

\begin{figure}[t]
  \centering
  \includegraphics[width=\linewidth]{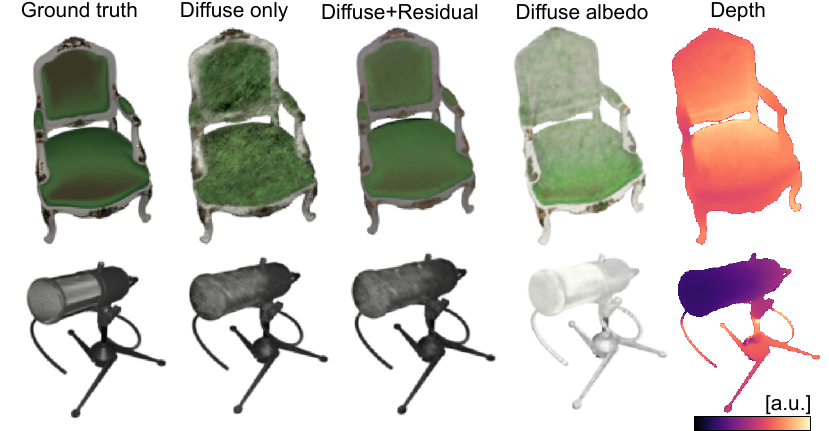}
  \caption{
    \label{fig:decomp}
    We visualize the decomposition of our model, as compared to the ground truth on two datasets. While the diffuse BSDF is unable to model the underlying BSDF, by combining it with a fully-learned component we are able to learn plausible reconstructions. The albedo is returned prior to any attenuation, and is thus much brighter than the output image.
  }
\end{figure}

\begin{figure*}
  \centering
  \includegraphics[width=\linewidth]{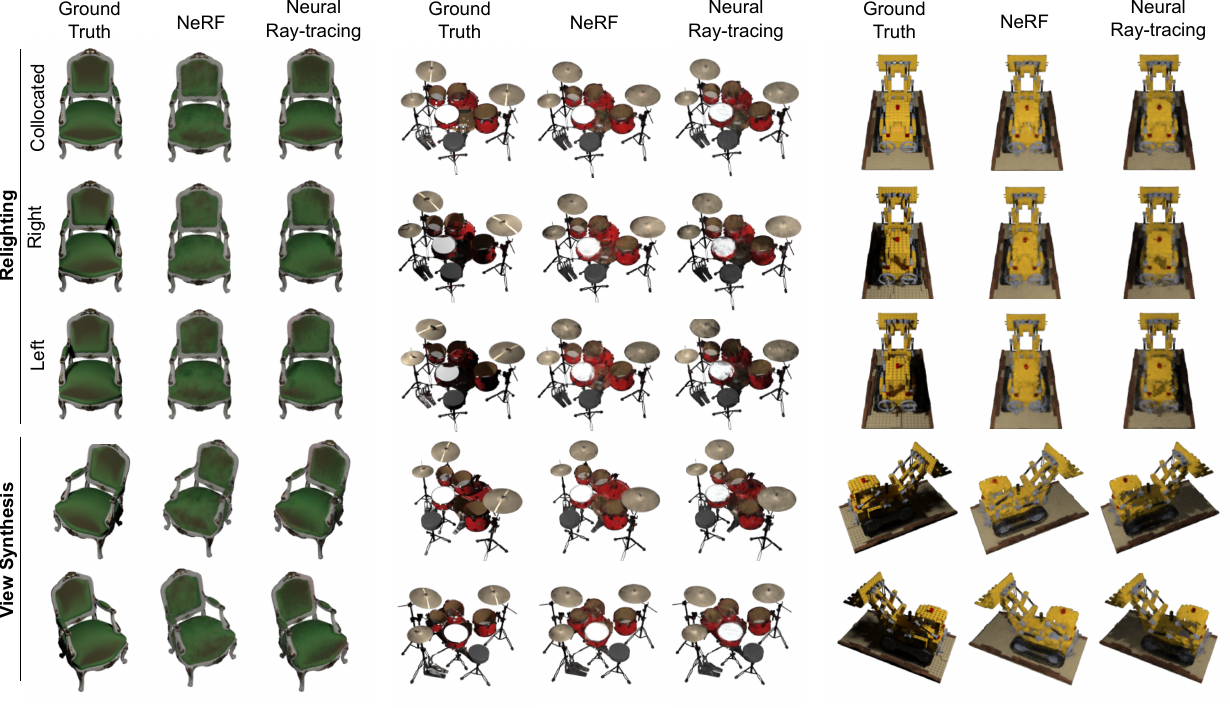}
  \caption{
    \label{fig:final}
    We qualitatively compare our method against NeRF~\cite{mildenhall2020nerf} on three
    synthetic scenes. Note the clean shadows produced by our method, especially in the Lego and
    Drums scenes, as well as the accurate appearance in the relit and novel view instances,
    especially in the velvet of the chair.
  }
\end{figure*}

\paragraph{Relighting and View-Synthesis}
With our estimated geometry and reflectance, we are fully equipped to tackle the applications of
relighting and view synthesis. We demonstrate relighting under novel conditions in
Figure.~\ref{fig:final}.
We test our models trained on the collocated light scenes on a test set of unseen 50 novel
illumination conditions and 50 novel views, synthesized by orbiting either the point light or
the camera around the scene while holding the other in place. Views in these test sets therefore
contain shadows, requiring our model to learn a reasonable occlusion model, despite seeing few
shadows in the training data. At test time, we compute occlusion using the learned occlusion
model and an additional raycasting check, taking advantage of the geometry of our model.
Our model is able to accurately recover shadows under a variety of
different lighting conditions, as well as free-view movement. Notably, the shadows are
consistent across different views, and change in reflectance match the ground truth.  In
Table~\ref{tab:nerf_cmp}, we report the quantiative performance of our model compared to 
NeRF~\cite{mildenhall2020nerf}, confirming the qualitative trends.

%
%
\begin{table}[t]
  \centering
	\resizebox{\linewidth}{!}{
  \begin{tabular}{c|c|c|c|c}
    Scene & \multicolumn{2}{c|}{Hotdog} & \multicolumn{2}{c}{Lego} \\
    \hline
    & PSNR$^\uparrow$ & MS-SSIM$^\uparrow$
    & PSNR$^\uparrow$ & MS-SSIM$^\uparrow$ \\
    \hline
    NeRF   & 22.75 & 0.75 & 18.08 & 0.66 \\
    Ours   & \cellcolor{yellow!30}25.89 & \cellcolor{yellow!30}0.87 & \cellcolor{yellow!30}22.69 & \cellcolor{yellow!30}0.85
    \vspace{2mm} \\
    Scene & \multicolumn{2}{c|}{Materials} & \multicolumn{2}{c}{Drums} \\
    \hline
    & PSNR$^\uparrow$ & MS-SSIM$^\uparrow$
    & PSNR$^\uparrow$ & MS-SSIM$^\uparrow$ \\
    \hline
    NeRF   & 17.95 & 0.67 & 20.73 & 0.77 \\
    Ours   & \cellcolor{yellow!30}19.14 & \cellcolor{yellow!30}0.72 & \cellcolor{yellow!30}22.50 & \cellcolor{yellow!30}0.83 \\
  \end{tabular}
	}
  \caption{
    \label{tab:nerf_cmp}
    We compare the proposed method against NeRF~\cite{mildenhall2020nerf} on our synthesized,
    relit test scenes, training on collocated point light data. In NeRF, we pass the light vector in
    addition to the view direction for a fairer comparison. Our method provides better
    relighting accuracy as compared to NeRF, which cannot explicitly model occlusion.
  }
\end{table}

\begin{table}[t]
  \centering
	\resizebox{\linewidth}{!}{
  \begin{tabular}{c|c|c|c|c}
    Scene & \multicolumn{2}{c|}{Armadillo} & \multicolumn{2}{c}{Hotdog} \\
    \hline
    & PSNR$^\uparrow$ & SSIM$^\uparrow$ & PSNR$^\uparrow$ & MS-SSIM$^\uparrow$ \\
    \hline
    NeRF+LE   & 20.35 & 0.883 & 19.96 & \cellcolor{yellow!30}0.868 \\
    NeRF+Env  & 19.60 & 0.874 & 19.94 & 0.863 \\
    Bi et al. & 22.35 & 0.894 & 23.74 & 0.862 \\
    NeRV,NVF  & 22.14 & \cellcolor{yellow!30}0.897 & 23.93 & 0.863 \\
    Ours      & \cellcolor{yellow!30}25.11 & 0.883 & \cellcolor{yellow!30}26.19 & \cellcolor{yellow!30}0.868  \\
  \end{tabular}
	}
  \caption{
    \label{tab:nerv_cmp}
    We compare results against recent neural rendering methods under known lighting conditions, see text, on the
    single point-light dataset from~\cite{srinivasan2020nerv}, with additional results in the Supplemental Document.
  }
\end{table}

\begin{figure}[t]
  \centering
  \includegraphics[width=\linewidth]{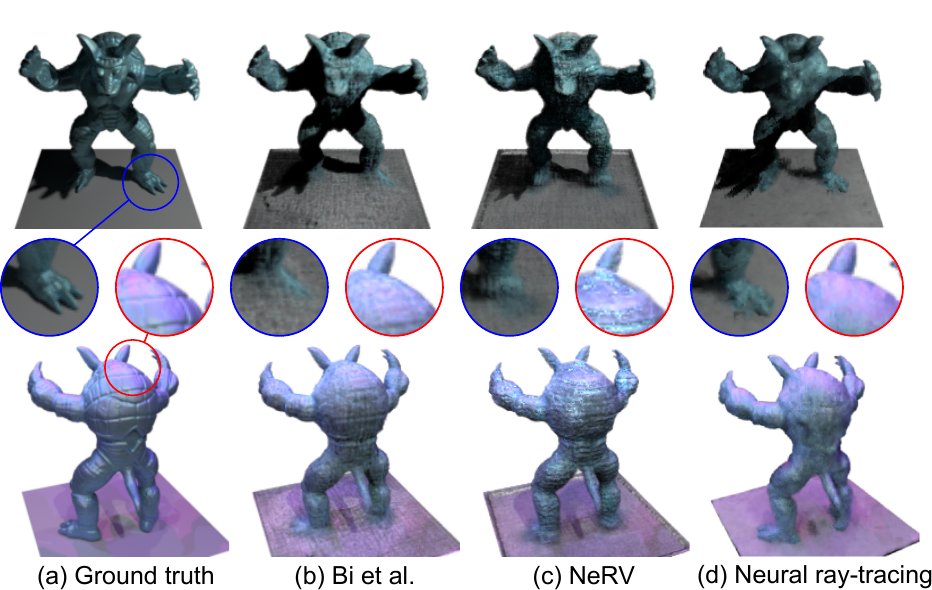}
  \caption{
    \label{fig:nerv_comparison_qual}
    Comparison of our neural ray tracing to Bi et al.~\cite{bi2020neural} and
    NeRV~\cite{srinivasan2020nerv}. We outperform these existing methods by reconstructing
    accurate details and appearance as shown in the insets. Notably, we more accurately capture
    the tone of the surface, as highlighted in the insets.
  }
\end{figure}

\paragraph{Comparison to Reflectance-Aware Methods}
We compare the proposed method to recent neural reflectance methods including NeRV~\cite{srinivasan2020nerv} and Bi et
al.~\cite{bi2020neural}. 
In addition, following~\cite{srinivasan2020nerv}, we also compare against a NeRF variant (noted as $\mathrm{NeRF+LE}$)
which combines the viewing direction with a light-embedding created with PointNet~\cite{qi2017pointnet}, to approximate varying light conditions. Note that we do not compare to NeRFactor since the scopes are different. We assume multiple, known lighting conditions for each view, whereas NeRFactor assumes a single, static, but unknown lighting condition. 
Figure~\ref{fig:nerv_comparison_qual} and Table~\ref{tab:nerv_cmp} show that our method outperforms existing methods both quantitatively and qualitatively, facilitated by our explicit surface model and smooth spatially-varying BSDF.
Bi et al.~\cite{bi2020neural} does not explicitly learn the true shadow, failing on accurate shadow reproducing.
It is also worth noting that our method is memory efficient training on a single GPU whereas  NeRV~\cite{srinivasan2020nerv} requires more computing resources, e.g., 128 TPUs in
their work. 
We note that the results of the compared methods are directly copied from the original papers of the authors as a public implementation was not made available (also confirmed in correspondence with the respective authors).


\paragraph{Disentangled Reflectance}
To validate the effectiveness of the proposed disentangled reflectance model, we evaluate novel view
synthesis for models with only diffuse Lambertian and fully-learned residual BSDFs.
Specifically, we simply modify Equation~\eqref{eq:bsdf} to output 1) only the diffuse component, 2) only the
learned component, and 3) both components together.
Figure~\ref{fig:decomp} and Table~\ref{tab:ablation} show the that the full disentangled reflectance model adds a significant increase in PSNR, but alone leads to much lower structural similarity.
The results validate that the combination of analytic and learned components offers higher representation power.

\begin{table}[t]
  \centering
	\resizebox{\linewidth}{!}{
  \begin{tabular}{c|c|c|c|c}
    Scene & \multicolumn{2}{c|}{Chair} & \multicolumn{2}{c}{Hotdog} \\
    \hline
    & PSNR$^\uparrow$ & MS-SSIM$^\uparrow$ & PSNR$^\uparrow$ & MS-SSIM$^\uparrow$ \\
    \hline
    Diffuse only    & 31.27 & 0.92 & 26.99 & 0.86  \\
    \hspace{-0.2em}Learned only\hspace{-0.2em}  & 27.74 & 0.77 & 25.63 & 0.81 \\
    Both                & \cellcolor{yellow!30}35.65 & \cellcolor{yellow!30}0.98 & \cellcolor{yellow!30}37.70 & \cellcolor{yellow!30}0.98
    \vspace{2mm}\\
    Scene & \multicolumn{2}{c|}{Lego} & \multicolumn{2}{c}{Drums} \\
    \hline
    & PSNR$^\uparrow$ & MS-SSIM$^\uparrow$ & PSNR$^\uparrow$ & MS-SSIM$^\uparrow$ \\
    \hline
    Diffuse only    & 20.51 & 0.71 & 19.60 & 0.61  \\
    \hspace{-0.2em}Learned only\hspace{-0.2em}  & 23.60 & 0.81 & 21.57 & 0.78  \\
    Both                & \cellcolor{yellow!30}29.43 & \cellcolor{yellow!30}0.95 & \cellcolor{yellow!30}23.92 & \cellcolor{yellow!30}0.90  \\
  \end{tabular}
	}
  \caption{
    \label{tab:ablation}
    Ablation study of our two-component model on multiple synthetic scenes. Rendering
    by diffuse color alone achieves comparatively high PSNR and MS-SSIM, but the learned-BSDF
    contribution further improves performance through its ability to model specular
    reflectance.
  }
\end{table}
\begin{figure}[t]
  \centering
  \includegraphics[width=\linewidth]{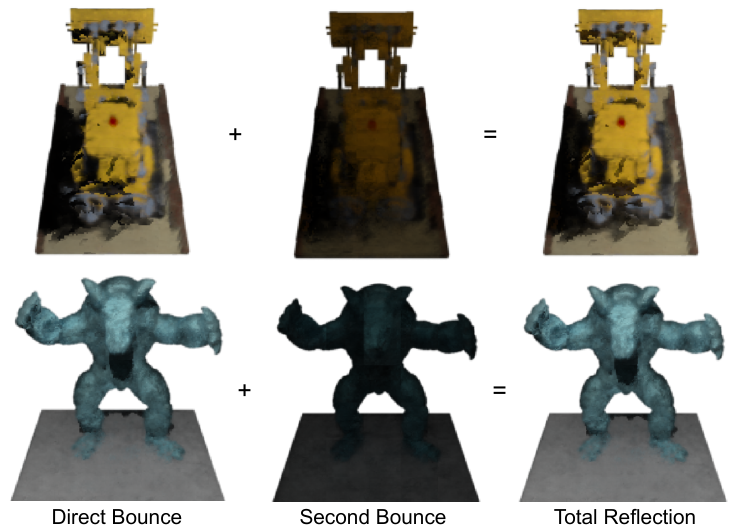}
  \caption{
    \label{fig:pathtrace_ord}
    Comparisons of the components of second-order pathtracing. The contribution from second order bounces provides indirect illumination in
    regions that are not directly visible from the scene illumination sources, see text.
  }
\end{figure}

\paragraph{Efficient Multi-Bounce Rendering with Pathtracing}
Our method efficiently performs multi-bounce rendering with path tracing.
Figure~\ref{fig:pathtrace_ord} shows the rendered images with direct reflection, second-bounce indirect reflection, and the total rendering which is a sum of the two components.
Our multi-bounce path tracing provides indirect illumination in the scene as compared to just the direct illumination.
Specifically, for the $\mathrm{Lego}$ scene, there is a significant amount of illumination on the roof, because the light is up and to the right of the Lego. The light reflects off the tractor claw onto the roof and vice-versa, and thus we see
significant illumination.
For the $\mathrm{Armadillo}$ scene, the light is collocated with the
camera, and we see indirect illumination on the face which may be reflection from other
components of the face which are strongly illuminated.
We achieve this multi-bounce rendering feature in a more efficient manner compared to NeRV~\cite{srinivasan2020nerv}, see Supplemental Material.
Figure~\ref{fig:nerv_comparison_pathtracing} describes the difference of our method to NeRV in multi-path rendering.
As described previously, we utilize bisectioning and raycasting to facilitate computation- and memory-efficient multi-bounce rendering.

\paragraph{Experimental Data with Unknown Lighting}\label{sec:dtu}
To test the applicability of the proposed method on real-world data, where the illumination of the scene is unknown and may
vary frame by frame, we recover views from the DTU dataset. The DTU image data which was captured experimentally in a illuminated
environment~\cite{jensen2014large}, and we do not use the ground truth illumination measurements.
To test whether the proposed method may be extended to unknown illumination scenarios, we modify our method and model spatially-varying lighting as
another neural scene representation $\mathrm{MLP}(\mathbf{x},\mathbf{z}_i)$ which takes
the scene position $\mathbf{x}\in\mathbb{R}^3$ and the per-image latent vector
$\mathbf{z}_i\in\mathbb{R}^c, c = 64$, which is optimized during training.
The lighting function approximates per-frame scene lighting by returning the intensity and
direction of the dominant light for each position.
We accurately
recover normals and depth as shown in Figure~\ref{fig:dtu}. For relighting, see Supplemental Document.
\begin{figure}[t]
  \centering
  \includegraphics[width=\linewidth]{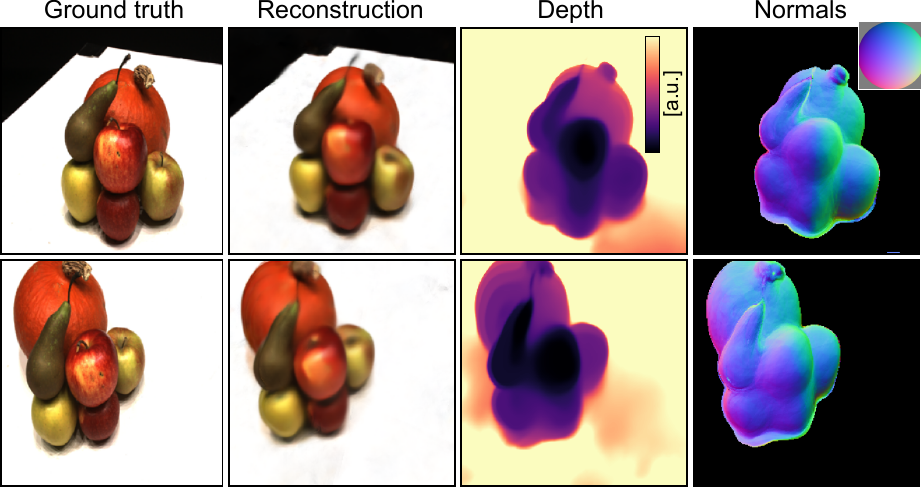}\\
  \caption{
    \label{fig:dtu}
    Experimental Reconstruction. When modified to account for unknown lighting, see text, the method is capable of handling
    experimentally-captured data with unknown lighting, recovering plausible depth and normals,
    due to the prior over the normals induced by the diffuse BSDF.
  }
\end{figure}

\section{Conclusion}
The proposed explicit modelling of analytic and learned components allows us to directly model light
transport, and allows for efficient surface intersections, permitting approximations of
soft-shadows, accurate surface reconstruction and fast higher-order light computations.
As such, our work bridges the gap between recent MLP-based approaches for neural rendering which rely on
a more decomposable pipeline and traditional physically-based rendering approaches. We show that
it is possible to combine concepts from conventional, physically-based, inverse-rendering
methods with high-fidelity scene representations via neural networks.  We hope future neural rendering methods to move
towards more explainable learned components, which are currently modelled as block box MLPs in a
large body of existing methods in neural rendering, making today's methods more inefficient,
uninterpretable, and challenging to integrate with the rich set of tools and methods developed
in traditional rendering approaches -- limitations that the proposed method makes steps towards
lifting.


{\small
  \bibliographystyle{ieee_fullname}
  \bibliography{ref}
}

\end{document}